\documentclass[runningheads]{llncs}
\usepackage{color}
\usepackage{graphicx}
\usepackage{geometry}
\title{Multi-Level Temporal Pyramid Network for Action Detection
\thanks{The first author of this paper is a graduate student.}}
\author{Xiang Wang\inst{1} \and Changxin Gao\inst{1} 
\and Shiwei Zhang\inst{2} \and Nong Sang\inst{1}\thanks{
Corresponding author. \quad
This work is supported by the National Natural Science Foundation of China under grant 61871435 and the Fundamental Research Funds for the Central Universities no.2019kfyXKJC024.
}
\\
\institute{Key Laboratory of Image Processing and Intelligent Control, \\
School of Artificial Intelligence and Automation, \\
Huazhong University of Science and Technology \and
DAMO Academy, Alibaba Group\\
\email{\tt\small \{u201613707, cgao, nsang\}@hust.edu.cn\\}
\email{\tt\small zhangjin.zsw@alibaba-inc.com}
}}
\begin{document}
\maketitle              
\begin{abstract}
Currently, one-stage frameworks have been widely applied for temporal action detection, but they still suffer from the challenge that the action instances
span a wide range of time. The reason is that these one-stage detectors, e.g., Single Shot Multi-Box Detector (SSD), extract temporal features only applying a single-level layer for each head, which is not discriminative enough to perform classification and regression. In this paper, we propose a Multi-Level Temporal Pyramid Network (MLTPN) to improve the discrimination of the features. Specially, we first fuse the features from multiple layers with different temporal resolutions, to encode
multi-layer temporal information. We then apply a multi-level feature pyramid architecture on the features to enhance their discriminative abilities. Finally, we design a simple yet effective feature fusion module to fuse the multi-level multi-scale features. By this means, the proposed MLTPN can learn rich and discriminative features for different action instances with different durations. We evaluate MLTPN on two challenging datasets: THUMOS’14 and Activitynet v1.3, and the experimental results show that MLTPN obtains competitive performance on Activitynet v1.3 and outperforms the state-of-the-art approaches on THUMOS’14 significantly.

\end{abstract}
\section{Introduction}
\begin{figure}[ht] 
    \includegraphics[width=\textwidth]{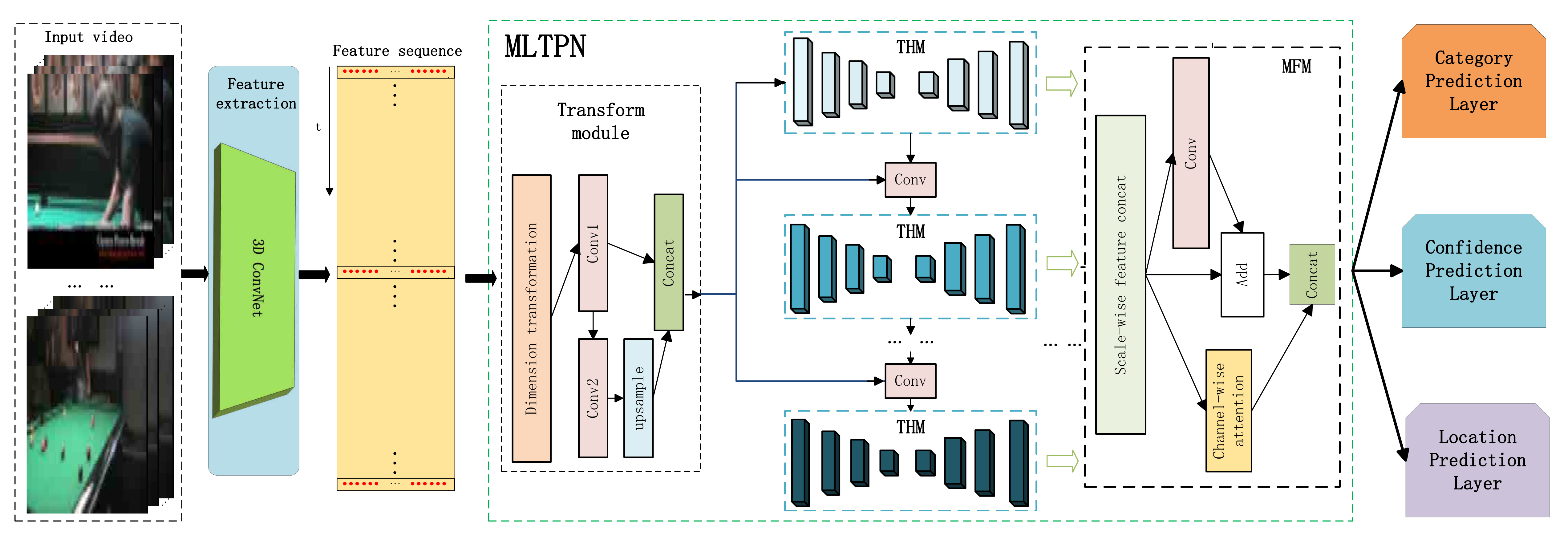}
    \caption{An overview of our MLTPN. We first utilize a 3D ConvNet to extract the features of the input video. Then the obtained feature sequence is input into Transform module to encode multi-layer temporal information. Afterwards
    each THM generates a group of multi-scale features, and then the cascaded multiple alternating joint Temporal H-shaped Modules (THMs) and Conv general multi-level pyramid features.
    Finally, Merge Feature Module (MFM) fuses the multi-level pyramid features for detection.}
   \label{fig:model}
\end{figure}

The purpose of temporal action detection in long untrimmed videos is to temporally localize intervals where actions occur and simultaneously predict the action categories.
It serves as a key technology in video retrieval, anomaly detection and human-machine interaction, hence it has been receiving an increasing mount of attention from both academia and industry.
Recently, temporal action detection has achieved great achievements on some public datasets~\cite{jiang2014thumos,Caba2015ActivityNet}.
However, this task is still challenging because the duration of action varies dramatically by ranging from fractions of a second to several minutes.
Recent existing approaches can be divided into three categories: multi-stage approach~\cite{lin2018bsn,lin2019bmn,li2019ctcn}, two-stage approach~\cite{xu2017rc3d,chao2018rethinking} and one-stage approach~\cite{lin2017ssad}.
Among these methods, one-stage methods, which are mainly inspired by SSD~\cite{liu2016ssd}, are more efficient and practical in many direct and indirect applications.
%
%
%
\begin{figure}   
    \centering
    \includegraphics[width=8cm]{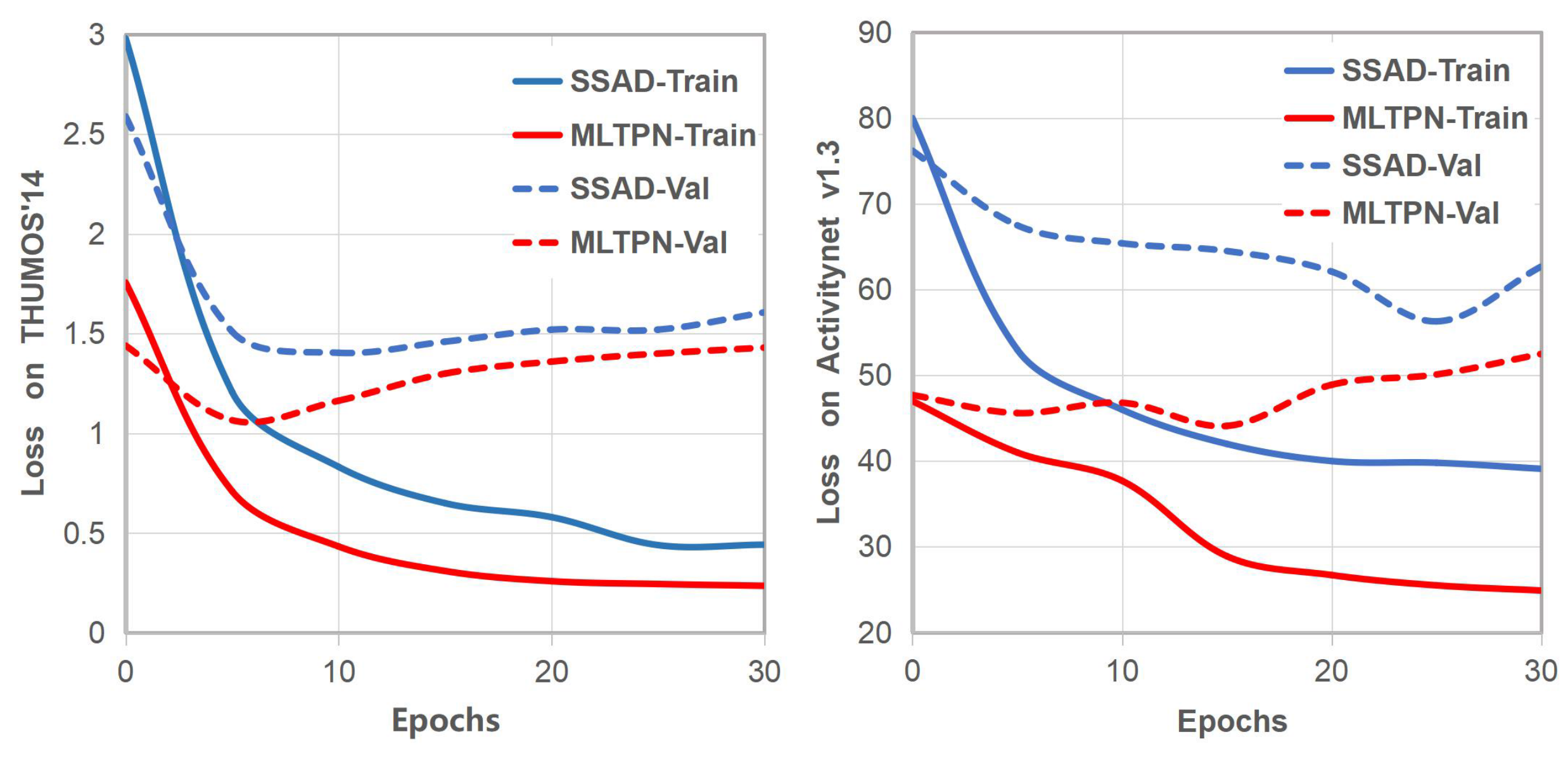}
    \caption{Losses of SSAD and MLTPN on the challenging THUMOS'14 (left) and Activitynet v1.3 (right) benchmarks. 
    Universally, SSAD losses are much higher than our MLTPN during both the training and validation phase. 
    The results demonstrate that MLTPN can encode more temporal information, and the encoded features are more discriminative to a certain degree. 
    Better viewed in original color pdf.}
   \label{fig:motivation}
\end{figure}
Typically, as with Single Shot Action Detector (SSAD) network~\cite{lin2017ssad}, these methods apply a single-level layer to detect actions for each head.
Because the action instances span a wide range of time, the SSD-like methods tend to apply the different layers to detect action with different durations, \emph{e.g.}, the shallow layers detect short actions and deep layers for long actions.
However, a single layer may be not discriminative enough to conduct action classification and localization, which can be proved in Figure~\ref{fig:motivation} to some extent.
Therefore, current performance of the methods are not satisfactory actually.
We believe that improving the discrimination of each layer can effectively improve action detection performance.
Based on the above observations, we proposed a Multi-Level Temporal Pyramid Network (MLTPN) to improve the performance of one-stage action detection, as is shown in Figure~\ref{fig:model}.
In our proposed MLTPN, we first encode multi-layer temporal information by fusing multiple layers with different temporal resolutions.
Second, inspired by the successful applications of multi-scale feature pyramid network~\cite{zhao2019m2det} in object detection, we propose to embed several feature pyramid networks on the encoded features to learn multi-level and discriminative features with different scales.
Third, we design a simple yet effective feature fusion module to fuse the multi-level multi-scale features.
By this means, the features encoded by the proposed MLTPN are more discriminative, and the undergoing multi-scale property is suitable for detection different action instances with different durations.
Moreover, to further improve the performance, we apply GIoU loss to regress the temporal boundaries.
To evaluate the effectiveness of the proposed MLTPN, we conduct experiments on THUMOS'14 and Activitynet v1.3 datasets, and the results show that MLTPN achieves a significant improvement on both datasets. Particularly, the result of MLTPN outperforms the state-of-the-art approaches on THUMOS’14 significantly.

In summary, we make the following three contributions:
\begin{itemize} 
    \item We propose a multi-level temporal feature pyramid network to improve the discrimination of each layer for one-stage frameworks;
    \item We design a simple but effective module to fuse multi-level multi-scale features;
    \item We extensively evaluate MLTPN on the challenging THUMOS'14 dataset and achieve state-of-the-art performance.
\end{itemize}
%
%
%
%
%
\section{Relation Work}
For temporal action detection task, recent existing approaches can be intuitively divided into three categories: multi-stage approach, two-stage approach and one-stage approach.\\
\indent Multi-stage approach,
particularly, Boundary Sensitive Network (BSN)~\cite{lin2018bsn} first locates temporal boundaries (start and end), then directly combines these boundaries as proposals. Next BSN retrieves proposals by evaluating the confidence of whether a proposal contains an action within its region. Finally, using the proposals to localize action.
Boundary-Matching Network (BMN)~\cite{lin2019bmn} is an improvement of BSN, first BMN predicts start and end boundary probabilities by a sub-network.
Then the boundary probabilities are applied to extensively enumerate the proposals, which is followed by a boundary-matching confidence map to densely evaluate confidence of all proposals.
Based on the proposals, then refine the boundaries and predict the corresponding categories. 
Multi-granularity Generator (MGG)~\cite{liu2019MGG} first uses a bilinear matching model to exploit the rich local information within the video. Then two components, namely segment proposal producer and frame actionness producer, are combined to perform the task of temporal at two distinct granularities. Finally, using the proposals to localize action.
These methods achieve impressive performance, but it is inefficient because of its long pipeline actually.
P-GCN~\cite{zeng2019graph} exploits the proposal-proposal relations using Graph Convolutional Networks, first based on the already obtained proposals, P-GCN constructs an action proposal graph, where each proposal is represented as a node and their relations between two proposals as an edge. Finally P-GCN applies the Graph Convolutional Networks over the graph to model the relations among different proposals and learn representations for the action classification and localization.\\
\indent As for two-stage approach, Region Convolutional 3D Network (R-C3D)~\cite{xu2017rc3d}, first encodes the video streams using a 3D fully convolutional network,then generates candidate temporal regions containing activities, and finally classifies selected regions into specific activities.
TAL-Net~\cite{chao2018rethinking} is an improvement of R-C3D, compared to R-C3D,TAL-net improves receptive field alignment, better exploits the temporal context of actions for both proposal generation and action classification by appropriately extending receptive fields.
Although, these methods have great improved on temporal action detection, 
The two-stage methods are Faster RCNN-like procedure, and suffer from another drawback that they are limited by fixed length inputs.
They need to down sample the frames to fit the GPU memory, \emph{e.g.}, 3 FPS is applied in~\cite{xu2017rc3d}.
Therefore, they lose some temporal information, which may result in a sub-optimal solution.\\
\indent In contrast, one-stage methods are mainly spirited by Single Shot MultiBox Detector (SSD), classification and localization at the same time, hence they are more efficient.  
SSAD~\cite{lin2017ssad} based on 1D temporal convolutional layers to skip the proposal generation step via directly detecting action instances in untrimmed videos. 
However, SSAD extracts temporal features only applying a single-level layer for each head, which is not discriminative enough to perform classification and localization. Therefore, we propose a Multi-Level Temporal Pyramid Network (MLTPN) to improve the discrimination of the features. In particular, SSAD is our baseline.
%
%
%
%
%
%
%
%
%
%
\section{The Proposed Method}
\subsection{Overview}
For the input video, we first use a 3D convolution network to extract features, and concatenate these features. Then the features are input into Transform module, to do the dimensional transformation and fuse the features from multiple layers. Afterwards, We apply a multi-level feature pyramid architecture on the features. Finally, the multi-level multi-scale features are input into the Merge Feature Module (MFM). MFM performs a certain fusion of the features, making our features more robust and more conducive. Finally, the features are used for action classification and localization.
\subsection{Feature Extracting}
To detect action instances in temporal dimension is the ultimate target of action localization. Given the video frame sequence, We uniformly sample the sequence of video frame into several small consecutive snippets and then extract visual features within each snippet. In particular, a sequence of snippets level feature $\{ {f_i}\} _{i = 0}^{ T - 1}$ are extracted, where $T$ is temporal length. We further feed the features into two 1D convolutional layers (with temporal kernel size 3, stride 2) to increase the temporal size of receptive fields.
%
%
\subsection{MLTPN}
Our MLTPN consists of three parts, namely transform module, Temporal H-Shaped Module (THM), Merge Feature Module (MFM).\\
%
%
\indent In the transform module, we use $t \times c$ to represent the obtained feature map, where $t$ is the temporal length and $c$ is the dimension of the representation. In particular, same as~\cite{li2019ctcn}, we first use a dimensional transformation, features change from $t \times c$ to  $1 \times t \times c$, and two convolutions to improve the dimension of features, so that our features contain multi-layer information and are more conducive to classification. 'Conv1' with $k1$ kernels, kernel size (3,1), stride (1,1). 'Conv2' with $k2$ kernels, kernel size (3,1), stride (2,1), and 'Conv2' also followed by one up-sample operation (Nearest neighbor interpolation or Linear interpolation) map back to the same size as output of 'Conv1' and we concatenate the outputs of 'Conv1' and 'Conv2'. The final output is $k \times t \times c$, where $k$ is equal to $k1 + k2$. Here, transform module can extract the multi-scare features from backbone.\\
\begin{figure}[ht]
    \centering
    \includegraphics[width=12cm]{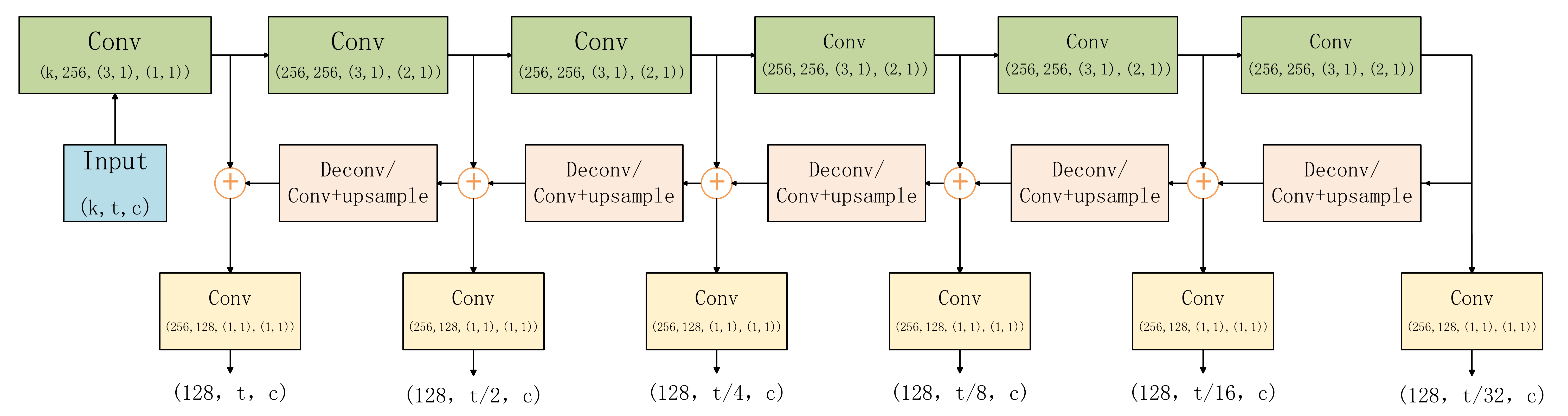}
    \caption{Details of THM. The numbers in brackets of Conv block:input channels, output channels, conv kernel size, stride size.}
  \label{fig:THM}
\end{figure}
\indent As shown in the Figure~\ref{fig:THM}, inspired by m2det~\cite{zhao2019m2det}, we design the THM. The encoder is a series of convolution layers (with kernel size (3,1), stride (2,1)). The output of these layers is reference set of feature map. At the same time, in order to make our feature representation more robust and more informative, we cascade multiple THMs. Finally the resulting multi-level multi-scale features, which are more conducive to subsequent tasks. 
The difference between cascaded multiple THMs and FPN (Feature Pyramid Network) includes several following aspects: 1) In FPN, each feature map in the pyramid is mainly or even solely constructed from single-level layers of the backbone. However, in our proposed cascaded multiple THMs, we embed several feature pyramid networks on the encoded features to learn multi-level and discriminative features with different scales; 2) The THM decoder takes the outputs of encoder layers as its reference set of feature maps, while the original FPN chooses the output of the last layer of each stage. 
Also the difference between cascaded multiple THMs and m2det is : In m2det, as the network deepens, the features of the model are reduced in both width and height, which is more suitable for object detection in the picture. However, our cascaded multiple THMs are reduced in one time dimension and the other is unchanged, which can well retain more semantic information suitable for classification.\\
\indent Conv between two adjacent THM consists of two convolution (with temporal kernel size 1, stride 1) for dimensionality reduction and a concat operation. \\
\indent The function of MFM is to fuse the features of different THM layers. First we concatenate the same scale features obtained by multiple THMs into $(l * 128) \times t' \times c$, where $l$ is the number of THM. The value of $t'$ is $t / 4$, $t / 8$, .. etc. Then we use a small residual module and a channel-wise attention module to fuse the multi-level features and the convolution in the small residual module shares parameters with different resolution features from the THM modules.
\subsection{Network Optimization}
For classification, we reshape the output of the category branch in each cell to a $(C+1)\times N$ matrix, $C+1$ represents the total action categories plus one background category. $N$ represents the number of anchor. The probability of being in the $i^{th}$ category is 
$$
\mathop P({C_i}) = \frac{{{e^{^{{o_i}}}}}}{{\sum\nolimits_{j = 1}^C {{e^{{o_j}}}} }},i = 0,1,2,...,C,\\
\eqno{(1)}
$$
where $o_i$ is the output of the network, which corresponds to the $i^{th}$ class. We utilize the standard softmax loss, which can be formulated as 
$$
\mathop {L_{cls}} =  - \sum\limits_{n = 0}^C {{I_{n = c}}\log (p({C_n}} ))\\
\eqno{(2)}
$$
where $I_{n = c}$ is an indicator function, equals to 1 if $n$ is the ground truth class label $c$, otherwise 0.\\
\indent We use the outputs of confidence branch as predictions of the IoU (Intersection over Union) values between the proposals and the ground truth. Note that the IoU here represents the temporal IoU, and there is only one temporal dimension. At the same time, for the sake of confusion, we will always use IoU to represent the temporal IoU. The IoU overlap loss is Smooth L1 loss ($S_{L1}$), so the loss is formulated as 
$$
\mathop {L_{conf}} = {S_{L1}}({p_{iou}} - {g_{iou}})\\
\eqno{(3)}
$$
where ${g_{iou}}$ is the ground truth IoU value between the proposal and its closest ground truth.\\
\indent For location prediction, the localization loss is only applied to positive samples. The same as ~\cite{rezatofighi2019giou}:
$$
\mathop {Io{u_i} = \frac{{\left| {{p_i} \cap {g_i}} \right|}}{{\left| {{p_i} \cup {g_i}} \right|}}} \\
\eqno{(4)}
$$
$$
\mathop {GIo{u_i} = Io{u_i} - \frac{{\left| {{B_i}\backslash ({p_i} \cup {g_i})} \right|}}{{\left| {{B_i}} \right|}}} \\
\eqno{(5)}
$$
$$
\mathop {L_{reg}} = 1 - GIo{u_i}  \\
\eqno{(6)}
$$
where ${g_i}$ is the closest ground truth of the proposal ${p_i}$. $B_i$ is the shortest continuous time interval including ground truth ${g_i}$ and the proposal ${p_i}$.\\
\indent We jointly train to optimize the above three loss functions. And the final loss is formulated as 
\small{
$$
\mathop {\begin{array}{c}
{L_{\cos t}} = {\alpha _1}\frac{1}{{{N_{cls}}}}\sum\limits_i {{L_{cls}}({p_i},p_i^*)}  + \\
{\alpha _2}\frac{1}{{{N_{conf}}}}\sum\limits_i {{L_{conf}}(Io{u_i},Io{u^*})}
+ {\alpha _3}\frac{1}{{{N_{reg}}}}\sum\limits_i {{L_{reg}}(GIo{u_i})} 
\end{array}} \\
\eqno{(7)}
$$}
where $\alpha _1$, $\alpha _2$ and $\alpha _3$ are weight coefficients (empirically set to 1, 10 and 0.3) . $N_{cls}$, $N_{conf}$ and $N_{reg}$ are the number of samples.
$p_i$ is the predicted probability and $p_i^*$ is the ground truth
label. $Io{u_i}$ is the predicted IoU between the sample and the closest ground truth , and $Io{u^*}$ is the real IoU. $GIo{u_i}$ is the $GIoU$ between the sample and the closest ground truth.
%
\section{Experiments}
We conduct experiments on two challenging datasets, THUMOS'14 and Activitynet v1.3 respectively. 
The impact of different experimental settings of MLTPN is investigated by
ablation studies. 
Comparison of MLTPN and other state-of-the-art methods is also reported.
%
\subsection{Dataset and Experimental Settings}
%
%
\noindent 
\textbf{Dataset. }
THUMOS’14 has 200 and 213 videos with temporal annotations in validation and testing set from 20 classes. The training set of THUMOS'14 is trimmed videos and cannot be used for training of temporal action detection for untrimmed videos. Therefore, like SSAD, we use its validation set to training our model and report results on its testing set. The Activitynet v1.3 dataset contains 19994 videos in 200 classes. The numbers of videos for training, validation and testing are 10024, 4926, and 5044. The labels of testing set are not publicly available and we report results on its validation set. \\
%
%
%
\noindent 
\textbf{Evaluation Metrics. }
We follow the official evaluation metrics in each dataset for action detection task. On THUMOS’14, the mean average precision (mAP) with IoU thresholds 0.3, 0.4, 0.5, 0.6, 0.7 are adopted. On Activitynet v1.3, the mAP with IoU thresholds between 0.5 and 0.95 (inclusive) with a step size 0.05 are exploited for comparison.\\
%
%
\noindent 
\textbf{Features. }
In order to extract snippet-level video feature, we first sample frames from each video at its own original frame rate on THUMOS’14 and 5fps on Activitynet v1.3. Then we apply a TV-L1~\cite{zach2007tv-l1} algorithm to get the optical flow of each frame. On THUMOS’14, we use open source I3D model~\cite{carreira2017I3D} pretrained on Kinetics to extract the I3D features,also we use the TSN models pretrained on Activitynet v1.3 to extract two-stream features. On activitynet v1.3, the pretrained two TSN models with Senet152~\cite{hu2018senet} are used to extract two-stream features. \\
%
%
\noindent 
\textbf{Implementation Details. }
For Thumos’14, In the training phase, we train the model using Stochastic Gradient Descent (SGD) with momentum of 0.9, weight decay of 0.0001 and the batch-size is set to 16. we set the initial learning rate at 0.001 and reduce once with a ratio of 0.1 after 15 epochs. For activitynet v1.3, we train the model using adam~\cite{kingma2014adam} method. we set the initial learning rate at 0.0001 and reduce once with a ratio of 0.1 after 15 epochs. The model is trained from scatch. Also we prevent model overfitting by using an early-stop strategy. In the testing phase, since there are little overlap between action instances of same category in
temporal action detection task, we take a strict threshold in NMS,
which is set to 0.2. \\
%
\begin{table}
\setlength{\tabcolsep}{1.1mm}
\parbox{.45\linewidth}{
\centering
\caption{Study of different feature settings on THUMOS'14 in terms of mAP@IoU(\%)}
\label{tab:booktabs1}
		\centering				
		\begin{tabular}{cccccc}
        \hline
			IoU	&	0.3	& 0.4 & 0.5 & 0.6 & 0.7	\\
			\hline
			TSN RGB	&	39.2 &	33.4 & 26.0 & 16.2 & 9.3 \\
			TSN FLOW &	46.6 &	43.1 & 35.0 & 24.5 & 13.4		\\
			TSN Fusion	& \textbf{57.2} & \textbf{51.7} & \textbf{42.5} &	\textbf{29.2} & \textbf{16.2}		\\
			\hline
			I3D RGB & 56.4 & 51.1 & 41.9 & 29.6 & 17.3    \\
			I3D FLOW & 62.6 & 58.0 & 49.4 & 34.9 & 20.0   \\
			I3D Fusion & \textbf{66.0} & \textbf{62.6} & \textbf{53.3} & \textbf{37.0} & \textbf{21.2}   \\
\hline
		\end{tabular}
}
\hfill
%
\parbox{.45\linewidth}{
\centering
\caption{Study of different anchor settings on THUMOS'14 in terms of mAP@IoU(\%)}			
\label{tab:booktabs2}
		\centering				
		\begin{tabular}{cccccc}
			\hline
			IoU	&	0.3	& 0.4 & 0.5 & 0.6 & 0.7	\\
			\hline
			5 anchors	&\textbf{66.3} &	61.3 & 50.0 & 35.1 & 18.9 \\
			6 anchors  & 63.8 &	59.1 & 50.0 & 33.2 & 19.1		\\
			7 anchors	& 66.0 & \textbf{62.6} & \textbf{53.3} &	\textbf{37.0} & \textbf{21.2}		\\
			8 anchors & 65.9 & 61.6 & 50.1 & 33.3 & 19.2\\
			9 anchors & 65.5 & 61.6 & 51.6 & 35.0 & 19.7\\
\hline
		\end{tabular}}
\end{table}
\begin{table}
\centering
\setlength{\tabcolsep}{1.1mm}
\caption{Study of different THM settings on THUMOS'14 in terms of mAP@IoU(\%)}			
\label{tab:booktabs3}
		\centering				
		\begin{tabular}{cccccc}
			\hline
			IoU	&	0.3	& 0.4 & 0.5 & 0.6 & 0.7	\\
			\hline
			4 THM	& 65.1 &59.8 & 50.4 & 35.8 & 20.4 \\
			5 THM  & 64.8 &	60.6 & 51.0 & 35.8 & \textbf{21.9}		\\
			6 THM	& \textbf{66.0} & \textbf{62.6} & \textbf{53.3} &	\textbf{37.0} & 21.2		\\
			7 THM & 63.8 & 59.2 & 48.8 & 34.2 & 19.3\\
			8 THM & 63.3 & 57.3 & 47.8 & 35.6 & \textbf{21.9}\\
			\hline
		\end{tabular}
\end{table}
\begin{table}[h]
\centering
\caption{Performance comparisons of temporal action detection on THUMOS'14, measured by mAP at different IoU thresholds. The contents of the brackets represent the feature extractor used.}
\label{tab:booktabs5}
\centering
\renewcommand\arraystretch{1.3}  
\setlength{\tabcolsep}{1.1mm}{
%
\scriptsize
\begin{tabular}{llllll}
\hline
\multicolumn{6}{c}{\textbf{THUMOS'14,mAP@IoU(\%)}} \\ \hline
\multicolumn{1}{c|}{Approach} & 0.3 & 0.4 & 0.5 & 0.6 & 0.7 \\ \hline
\hline
\multicolumn{6}{c}{Multi-stage\&Two-stage Action Localization} \\ \hline
\multicolumn{1}{c|}{\footnotesize{Wang et al.~\cite{wang2014action}}} & 14.0 & 11.7 & 8.3 & - & - \\
\multicolumn{1}{c|}{\footnotesize{FTP~\cite{caba2016fast}}} & - & - & 13.5 & - & - \\
\multicolumn{1}{c|}{\footnotesize{DAP~\cite{escorcia2016daps}}} & - & - & 13.9 & - & - \\
\multicolumn{1}{c|}{\footnotesize{Oneata et al.~\cite{oneata2014lear}}} & 27.0 & 20.8 & 14.4 & - & - \\
\multicolumn{1}{c|}{\footnotesize{Yuan et al.~\cite{yuan2016temporal}}} & 33.6 & 26.1 & 18.8 & - & - \\
\multicolumn{1}{c|}{\footnotesize{SCNN~\cite{shou2016sscn}}} & 36.3 & 28.7 & 19.0 & 10.3 & 5.3 \\
\multicolumn{1}{c|}{\footnotesize{SST~\cite{buch2017sst}}} & 41.2 & 31.5  & 20.0 & 10.9 & 4.7 \\
\multicolumn{1}{c|}{\footnotesize{CDC~\cite{shou2017cdc}}} &  40.1 & 29.4 & 23.3 & 13.1 & 7.9  \\
\multicolumn{1}{c|}{\footnotesize{TURN~\cite{gao2017turn}}} &  46.3 & 35.5 & 24.5 & 14.1 & 6.3  \\
\multicolumn{1}{c|}{\footnotesize{TCN~\cite{dai2017TCN}}} &- & 33.3 & 25.6 & 15.9 & 9.0  \\
\multicolumn{1}{c|}{\footnotesize{R-C3D~\cite{xu2017rc3d}}} &  44.8 & 35.6 & 28.9 & 19.1 & 9.3  \\
\multicolumn{1}{c|}{\footnotesize{SSN~\cite{zhao2017SSn}}} &  51.9 & 41.0 & 29.8 & 19.6 & 10.7  \\
\multicolumn{1}{c|}{\footnotesize{CBR~\cite{gao2017cascaded}}} &  50.1 & 41.3 & 31.0 & 19.1 & 9.9  \\
\multicolumn{1}{c|}{\footnotesize{BSN~\cite{lin2018bsn}}} &  53.5 & 45.0 & 36.9 & 28.4 & 20.0  \\
\multicolumn{1}{c|}{\footnotesize{MGG~\cite{liu2019MGG}}} & 53.9 & 46.8 & 37.4 & 29.5 & \textbf{21.3}  \\
\multicolumn{1}{c|}{\footnotesize{BMN~\cite{lin2019bmn}}} & 56.0 & 47.4 & 38.8 & 29.7 & 20.5 \\
\multicolumn{1}{c|}{\footnotesize{TAL-Net~\cite{chao2018rethinking}}} &   53.2 & 48.5 & 42.8 & 33.8 & 20.8  \\
\multicolumn{1}{c|}{\footnotesize{P-GCN~\cite{zeng2019graph}}} &  63.6 & 57.8 & 49.1 & - & -  \\ \hline
\hline
\multicolumn{6}{c}{One-stage Action Localization} \\ \hline
\multicolumn{1}{c|}{\footnotesize{Richard et al.~\cite{richard2016temporal}}} &  30.0 & 23.2 & 15.2 & - & -  \\
\multicolumn{1}{c|}{\footnotesize{Yeung et al.~\cite{yeung2016end}}} &  36.0 & 26.4 & 17.1 & - & -  \\
\multicolumn{1}{c|}{\footnotesize{SMS~\cite{yuan2017SMS}}} &36.5 & 27.8 & 17.8 & - & - \\
\multicolumn{1}{c|}{\footnotesize{SS-TAD~\cite{buch2017sstad}}} & 45.7 & - & 29.2 & - & 9.6 \\
\multicolumn{1}{c|}{\footnotesize{SSAD~\cite{lin2017ssad} (TSN)}} & 52.6 & 46.2 & 36.6 & 22.8 & 12.0 \\
\multicolumn{1}{c|}{\footnotesize{SSAD~\cite{lin2017ssad} (I3D)}} & 62.53 & 55.14 & 42.1 & 27.43 & 13.5 \\
\hline
\multicolumn{1}{c|}{\footnotesize{\textbf{MLTPN} (TSN)}}& 57.2 & 51.7 & 42.5 & 29.2 & 16.2 \\ 
\multicolumn{1}{c|}{\footnotesize{\textbf{MLTPN} (I3D)}}& \textbf{66.0} & \textbf{62.6} & \textbf{53.3} & \textbf{37.0} & 21.2 \\ 
\hline
\end{tabular}}
\end{table}
\subsection{Ablation Studies}
%
%
%
\textbf{Feature Settings. }
We use the features generated on the TSN and I3D models for comparison experiments, as shown in Table~\ref{tab:booktabs1}. It can be seen that the I3D feature performance is much better than TSN features, and there is a large gap. Moreover, it is common that FLOW features perform better than RGB features. As expected, the fusion features perform better than using only the RGB or FLOW features, indicating that in this task, the RGB features and the FLOW features have a certain complementary effect.\\
%
%
%
\begin{table}[ht]
\caption{Performance comparisons of temporal action detection on Activitynet v1.3. ($*$) indicates the method that uses the external video labels from ~\cite{zhaoy}}
\label{tab:booktabs7}
\footnotesize
\renewcommand\arraystretch{1.2}   
\centering
\setlength{\tabcolsep}{1.4mm}{   
\begin{tabular}{lllll}
\hline
\multicolumn{5}{c}{\textbf{Activitynet v1.3@IoU(\%)}} \\ \hline
\multicolumn{1}{c|}{Approach} & \multicolumn{4}{c}{validation} \\ \cline{2-5}  
\multicolumn{1}{c|}{} & 0.5 & 0.75 & 0.95 & \footnotesize{Average} \\ \hline

\hline
\multicolumn{5}{c}{Multi-stage\&Two-stage Action Localization} \\ \hline
\multicolumn{1}{c|}{\footnotesize{Wang et al.~\cite{wang2016uts1}}} & 45.11 & 4.11 & 0.05 & 16.41 \\
\multicolumn{1}{c|}{\footnotesize{CDC~\cite{shou2017cdc}}} & 45.30 & 26.00 & 0.20 & 23.80 \\
\multicolumn{1}{c|}{\footnotesize{TAG-D~\cite{xiong2017TAG}}} & 39.12 & 23.48 & 5.49 & 23.98 \\
\multicolumn{1}{c|}{\footnotesize{TAL-Net~\cite{chao2018rethinking}}} & 38.23 & 18.30 & 1.30 & 20.22 \\
\multicolumn{1}{c|}{\footnotesize{BSN$*$~\cite{lin2018bsn}}} & 46.45 & 29.96 & 8.02 & 30.03 \\
\multicolumn{1}{c|}{\footnotesize{BMN$*$~\cite{lin2019bmn}}} & 50.07 & 34.78 & 8.29 & 33.85 \\
\multicolumn{1}{c|}{\footnotesize{P-GCN~\cite{zeng2019graph}}} & 42.90 & 28.14 & 2.47 & 26.99 \\
\multicolumn{1}{c|}{\footnotesize{P-GCN$*$~\cite{zeng2019graph}}} & 48.26 & 33.16 & 3.27 & 31.11 \\ \hline

\multicolumn{5}{c}{One-stage Action Localization} \\ \hline
\multicolumn{1}{c|}{\footnotesize{Singh et al.~\cite{singh2016multi2}}} & 26.01 & 15.22 & 2.61 & 14.62 \\
\multicolumn{1}{c|}{\footnotesize{SSAD~\cite{lin2017ssad}}} & 33.91 & 23.16 & 3.74 & 22.26 \\
\multicolumn{1}{c|}{\footnotesize{\textbf{MLTPN}}} & \textbf{44.86} & \textbf{28.96} & \textbf{4.30} & \textbf{28.27} \\\hline
\end{tabular}}
\end{table}
\begin{figure} 
    \centering
    \includegraphics[width=\textwidth]{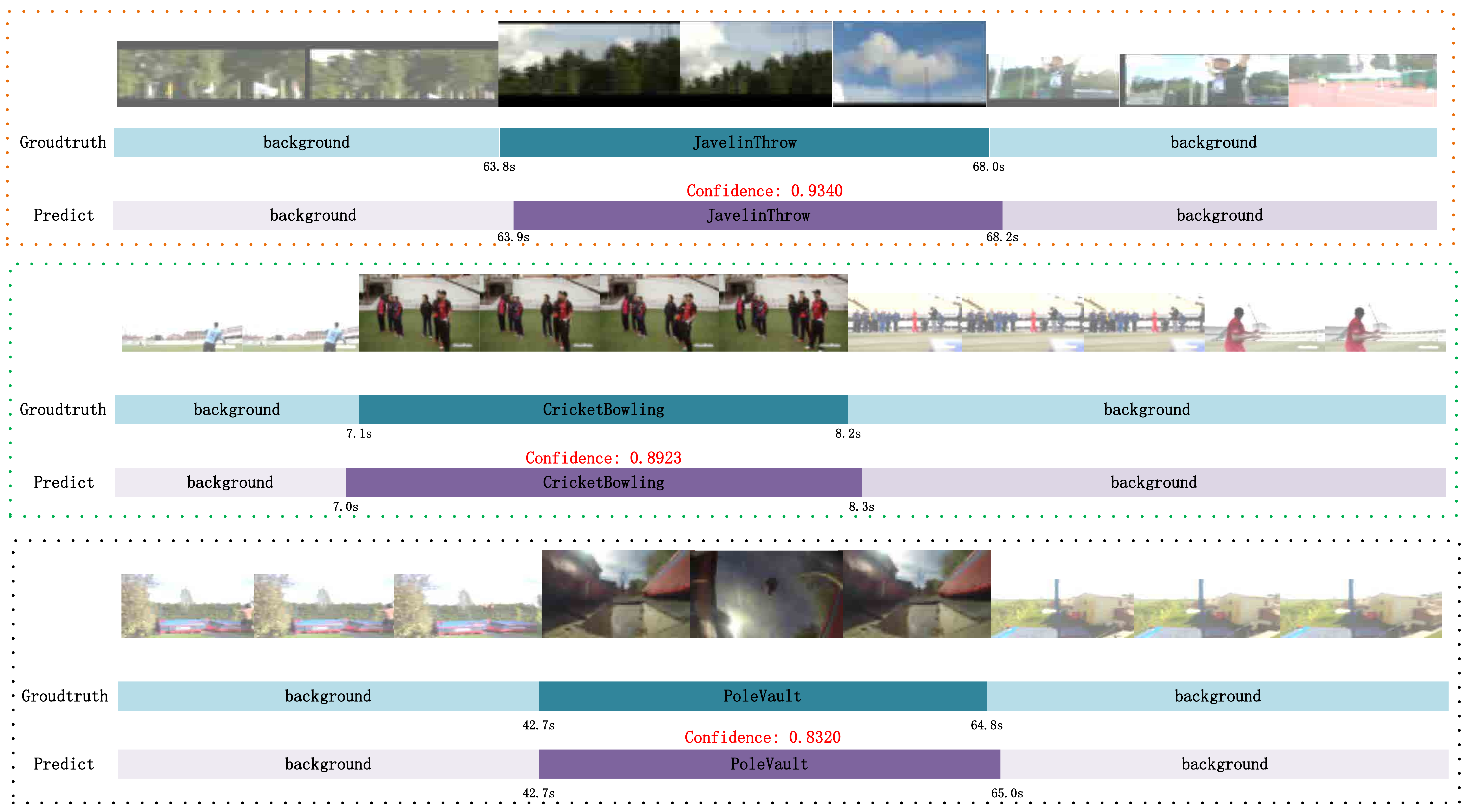}
    \caption{Visualization of predicted action instances by our MLTPN on THUMOS'14.}
  \label{fig:gt_pred}
\end{figure}
\noindent 
\textbf{Anchor Settings. }
We study the effect of different number of anchor segments on our performance. As shown in Table~\ref{tab:booktabs2}, the setting of 7 anchors achieved the best results with the IoU thresholds of 0.4, 0.5, 0.6, 0.7. The best results are obtained with the setting of 5 anchors with the IoU threshold of 0.3. We can say that the number of anchors cannot be too much, because many redundant detections are involved, and it also cannot be too few. Too few, there are too few examples of training, which is not conducive to training a good result.\\
%
\noindent 
\textbf{Number of THM. }
We study the effect of the different number of THM modules in our MLTPN. The results are shown in the Table~\ref{tab:booktabs3}. We find that when the number of THM modules is 6, on the THUMOS'14 dataset, the performance is best with the IoU thresholds of 0.3, 0.4, 0.5, 0.6, and when the number of THM modules is 5 and 8, the performance is best with the IoU threshold of 0.7.
\subsection{Comparison with the State-of-the-arts}
%
%
\textbf{THUMOS'14. }
We compare our MLTPN with some current state-of-the-art methods on THUMOS'14 and summarize the results in Table~\ref{tab:booktabs5}. we can see that our method has a great improvement over the current methods. Especially at IoU 0.3, 0.4, 0.5, 0.6, the performance is even better than the multi-stage methods and two-stage methods, achieving the current best results and it is very close to the best results at IoU 0.7. It is worth noting that the performance of MLTPN is much better than that of SSAD when MLTPN and SSAD use the same feature extractor (e.g. TSN, I3D). This fully proves the effectiveness of our method. And Figure~\ref{fig:gt_pred} shows the visualization of prediction results of three action categories respectively.\\
%
%
\noindent 
\textbf{Activitynet v1.3. }
On Activitynet v1.3 dataset, we compare with some current state-of-the-art methods. As shown in the Table~\ref{tab:booktabs7}, regarding the average mAP, MLTPN outperforms SSAD by 6.01\% with the same features and the same experimental Settings for fair comparison. Please refer to section 4.1 for experimental Settings. In fact, MLTPN has better performance than many current multi-stage \& two-stage methods (i.g., CDC, TAL-Net) and our method can make direct inference to the detection results without using any other labels, while many other multi-stage methods require step-by-step inference, which is very troublesome and time-consuming. For example, in the BMN approach, the classification results of the proposals must be obtained before they can be combined with the final detection results. And under the same experimental environment configuration, the calculation speed of BMN is 0.363 seconds per video (excluding classification), and MLTPN is 0.284 seconds per video (including classification).
\section{Conclusion}
In this paper, we proposed a Multi-Level Temporal Pyramid Network (MLTPN) for action detection. Our MLTPN drops the proposal generation step and can directly predict action instances in untrimmed videos. Specially, we first fuse the features from multiple layers with different temporal resolutions, to encode multi-layer temporal information. We then apply a multi-level feature pyramid architecture on the feature to enhance their discriminative abilities. Finally, we design a simple yet effective feature fusion module to fuse the multi-level multi-scale feature. Our MLTPN can learn rich and discriminative features for different action instances with different durations. As the experimental results show, MLTPN obtains competitive performance on Activitynet v1.3 and outperforms the state-of-the-art approaches on THUMOS'14 significantly.\\
\\

\end{document}